\documentclass[lettersize,journal]{IEEEtran}
\usepackage{amsmath,amsfonts}
\usepackage{algorithmic}
\usepackage{algorithm}
\usepackage{array}
\usepackage[caption=false,font=normalsize,labelfont=sf,textfont=sf]{subfig}
\usepackage{textcomp}
\usepackage{stfloats}
\usepackage{url}
\usepackage{verbatim}
\usepackage{graphicx}
\usepackage{cite}
\usepackage[hidelinks]{hyperref}
\usepackage{url}
\usepackage{pifont}
\usepackage{cuted}
\usepackage{xcolor}

\def\x{{\mathbf x}}														
\def\y{{\mathbf y}}														
\def\W{{\mathbf W}}														
\def\ii{{\hat{\imath}}}												
\def\ij{{\hat{\jmath}}}												
\def\ik{{\hat{\kappa}}}												
\def\bH{\mathbb{H}}														
\def\bR{\mathbb{R}}														

\begin{document}

\title{PHNNs: Lightweight Neural Networks via Parameterized Hypercomplex Convolutions}

\author{Eleonora Grassucci,~\IEEEmembership{Graduate Student Member,~IEEE,} Aston Zhang, and Danilo~Comminiello,~\IEEEmembership{Senior~Member,~IEEE}%
\thanks{E. Grassucci and D. Comminiello are with the Dept. Information Engineering, Electronics and Telecommunications (DIET), Sapienza University of Rome, Italy. A. Zhang is with Amazon Web Services AI, East Palo Alto, CA, USA. Corresponding author's email: eleonora.grassucci@uniroma1.it.}}



\maketitle

\begin{abstract}
Hypercomplex neural networks have proven to reduce the overall number of parameters while ensuring valuable performance by leveraging the properties of Clifford algebras. Recently, hypercomplex linear layers have been further improved by involving efficient parameterized Kronecker products. In this paper, we define the parameterization of hypercomplex convolutional layers and introduce the family of parameterized hypercomplex neural networks (PHNNs) that are lightweight and efficient large-scale models. Our method grasps the convolution rules and the filter organization directly from data without requiring a rigidly predefined domain structure to follow. PHNNs are flexible to operate in any user-defined or tuned domain, from 1D to $n$D regardless of whether the algebra rules are preset.
Such a malleability allows processing multidimensional inputs in their natural domain without annexing further dimensions, as done, instead, in quaternion neural networks for 3D inputs like color images.
As a result, the proposed family of PHNNs operates with $1/n$ free parameters as regards its analog in the real domain. We demonstrate the versatility of this approach to multiple domains of application by performing experiments on various image datasets as well as audio datasets in which our method outperforms real and quaternion-valued counterparts. Full code is available at: \url{https://github.com/eleGAN23/HyperNets}.
\end{abstract}

\begin{IEEEkeywords}
Hypercomplex Neural Networks, Kronecker Decomposition, Lightweight Neural Networks, Quaternions, Efficient models
\end{IEEEkeywords}

\section{Introduction}
\label{sec:intro}

\IEEEPARstart{R}{ecent} state-of-the-art convolutional models achieved astonishing results in various fields of application by large-scaling the overall parameters amount \cite{karras2020analyzing, dascoli2021convit, dosovitskiy2021image, Real2019ImgClass}. Simultaneously, hypercomplex algebra applications are gaining increasing attention in diverse spheres of research such as signal processing \cite{NAVARROMORENO2021108022, NAVARROMORENO202010100, Sanei2018ICASSP, XIANG2018193} or deep learning \cite{Kamayashi2021TNNLS, Lin2021TNNLS, Liu2021TNNLS, Valle2018TNNLS, Liu2018TNNLS, VALLE2020136, DECASTRO202054, PaulTNNLS2015, Hirose2014SSTNNLS}. Indeed, hypercomplex and quaternion neural networks (QNNs) demonstrated to significantly reduce the number of parameters while still obtaining comparable performance \cite{Muppidi2021ICASSP, ParcolletICLR2019, GrassucciQGAN2021, Tay2019QTRansformer, Cariow2021Oct, WU2020179, VALLE2021111}. These models exploit hypercomplex algebra properties, including the Hamilton product, to painstakingly design interactions among the imaginary units, thus involving $1/4$ or $1/8$ of free parameters with respect to real-valued models. Furthermore, thanks to the modelled interactions, hypercomplex networks capture internal latent relations in multidimensional inputs and preserve pre-existing correlations among input dimensions \cite{Chen2021QFM, GrassucciICASSP2021, Grassucci2021Entropy, Gai2021TCS, Vieira2020IJCNN}. Therefore, the quaternion domain is particularly appropriate for processing $3$D or $4$D data, such as color images or (up to) $4$-channel signals \cite{Took2019ICASSP}, while the octonion one is suitable for $8$D inputs. Unfortunately, most common color image datasets contain RGB images and some tricks are required to process this data type with QNNs. Among them, the most employed are padding a zero channel to the input in order to encapsulate the image in the four quaternion components, or remodelling the QNN layer with the help of vector maps \cite{Gaudet2021RemDim}. Additionally, while quaternion neural operations are widespread and easy to be integrated in pre-existing models, very few attempts have been made to extend models to different domain orders. Accordingly, the development of hypercomplex convolutional models for larger multidimensional inputs, such as magnitudes and phases of multichannel audio signals or $16$-band satellite images, still remains painful. Moreover, despite the significantly lower number of parameters, these models are often slightly slow with respect to real-valued baselines \cite{hoffmann2020algebranets} and ad-hoc algorithms may be necessary to improve efficiency \cite{Cariow2021Quat, Cariow2021Oct}.

Recently, a novel literature branch aims at compress neural networks leveraging Kronecker product decomposition \cite{Huang2020StochasticNN, Tang2021SKFAC}, gaining considerable results in terms of model efficiency \cite{Wang2021KroneckerCD}. Lately, a parameterization of hypercomplex multiplications have been proposed to generalize hypercomplex fully connected layers by sum of Kronecker products \cite{Zhang2021PHM}. The latter method obtains high performance in various natural language processing tasks by also reducing the number of overall parameters. Other works extended this approach to graph neural networks \cite{le2021parameterized} and transfer learning \cite{mahabadi2021compacter}, proving the effectiveness of Kronecker product decomposition for hypercomplex operations. However, no solution exists for convolutional layers yet, which remain the most employed layers when dealing with multidimensional inputs, such as images and audio signals \cite{Wu2021CvTIC, Hersheyicassp2017}.

In this paper, we devise the family of parameterized hypercomplex neural networks (PHNNs), which are lightweight large-scale hypercomplex neural models admitting any multidimensional input, whichever the number of dimensions. At the core of this novel set of models, we propose the parameterized hypercomplex convolutional (PHC) layer. Our method is flexible to operate in domains from $1$D to $n$D, where $n$ can be arbitrarily chosen by the user or tuned to let the model performance lead to the most appropriate domain for the given input data. Such a malleability comes from the ability of the proposed approach to subsume algebra rules to perform convolution regardless of whether these regulations are preset or not. Thus, neural models endowed with our approach adopt $1/n$ of free parameters with respect to their real-valued counterparts, and the amount of parameter reduction is a user choice. This makes PHNNs adaptable to a plethora of applications in which saving storage memory can be a crucial aspect.
Additionally, PHNNs versatility allows processing multidimensional data in its natural domain by simply setting the dimensional hyperparameter $n$. For instance, color images can be analyzed in their RGB domain by setting $n=3$ without adding any useless information, contrary to standard processing for quaternion networks with the padded zero-channel. Indeed, PHC layers are able to grasp the proper algebra from input data, while capturing internal correlations among the image channels and saving $66\%$ of free parameters.

On a thorough empirical evaluation on multiple benchmarks, we demonstrate the flexibility of our method that can be adopted in different domains of applications, from images to audio signals.
We devise a set of PHNNs for large-scale image classification and sound event detection tasks, letting them operate in different hypercomplex domain and with various input dimensionality with $n$ ranging from $2$ to $16$.

The contribution of this paper is three-fold.
\begin{itemize}
    \item We introduce a parameterized hypercomplex convolutional (PHC) layer which grasps the convolution rules directly from data via backpropagation exploiting the Kronecker product properties, thus reducing the number of free parameters to $1/n$.
    \item We devise the family of parameterized hypercomplex neural networks (PHNNs), lightweight and more efficient large-scale hypercomplex models. Thanks to the proposed PHC layer and to the method in \cite{Zhang2021PHM} for fully connected layers, PHNNs can be employed with any kind of input and pre-existing neural models. To show the latter, we redefine common ResNets, VGGs and Sound Event Detection networks (SEDnets), operating in any user-defined domain just by choosing the hyperparameter $n$, which also drives the number of convolutional filters.
    \item We show how the proposed approach can be employed with any kind of multidimensional data by easily changing the hyperparameter $n$. Indeed, by setting $n=3$ a PHNN can process RGB images in their natural domain, while leveraging the properties of hypercomplex algebras, allowing parameter sharing inside the layers and leading to a parameter reduction to $1/3$. To the best of our knowledge, this is the first approach that processes color images with hypercomplex-based neural models without adding any padding channel. As well, multichannel audio signals can be analysed by simply considering $n=4$ for standard first-order ambisonics (which has $4$ microphone capsules), $n=8$ for an array of two ambisonics microphones, or even $n=16$ if we want to include the information of each channel phase.

\end{itemize}

The rest of the paper is organized as follows. In Section \ref{subsec:real_layers}, we introduce concepts of hypercomplex algebra and we recapitulate real and quaternion-valued convolutional layers. Section \ref{sec:phc} rigorously introduces the theoretical aspects of the proposed method. Sections \ref{sec:phc_rgb} and \ref{sec:phc_audio} reveal how the approach can be adopted in different neural models and in two different domains, the images and audio one, expounding how to process RGB images with $n=3$ and multichannel audio with $n$ up to $8$. The experimental evaluation is presented in Section \ref{sec:img_class} for image classification and in Section \ref{sec:sed} for sound event detection. Finally, Section \ref{sec:abl} reports the ablation studies we conduct and in Section \ref{sec:conc} we draw conclusions.

\section{Hypercomplex Neural Networks}
\label{subsec:real_layers}

\subsection{Hypercomplex Algebra}

\begin{figure}[t]
    \centering
    \includegraphics[width=\linewidth]{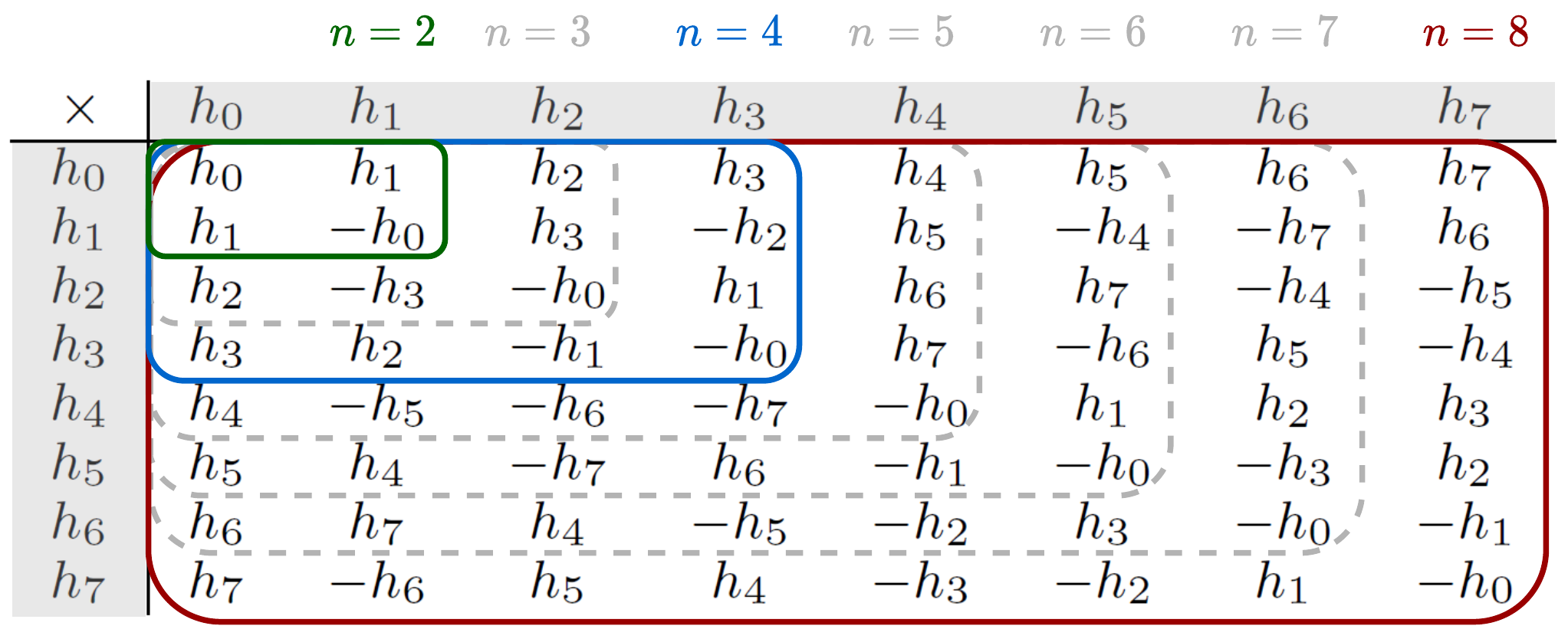}
    \caption{Example of hypercomplex multiplication table for $n=2$ i.e., complex, among others (green line), $n=4$ i.e., quaternions, tessarines, (blue line) and $n=8$, i.e., octonions, bi-quaternions, and so on (red line). While for these domains algebra rules exist and are predefined, no regulations are set for other domains such as $n=3,5,6,7$ (dashed grey lines). The parameterized hypercomplex approaches are able to learn these missing algebra rules from data, thus defining hypercomplex multiplication and convolution for any desired domain.}
    \label{fig:hprod}
\end{figure}

Hypercomplex neural networks rely in a hypercomplex number system based on the set of hypercomplex numbers $\bH$ and their corresponding algebra rules to shape additions and multiplications \cite{VALLE2021111}. These operations should be carefully modelled due to the interactions among imaginary units that may not behave as real-valued numbers. For instance,  Figure~\ref{fig:hprod} reports an example of a multiplication table for complex (green), quaternion (blue) and octonion (red) numbers. However, this is just a small subset of the hypercomplex domain that exist. Indeed, for $n=4$ there exist quaternions, tessarines, among others, while for $n=8$ octonions, dual-quaternions, and so on. Each of these domains have different multiplication rules due to dissimilar imaginary units interactions. A generic hypercomplex number is defined as

\begin{equation}
    h = h_0 + h_i \ii_i + \ldots + h_n \ii_n, \qquad i=1, \ldots, n
\label{eq:hyp_num}
\end{equation}

\noindent being $h_0, \ldots, h_n \in \bR$ and $\ii_i, \ldots, \ii_n$ imaginary units. Different subsets of the hypercomplex domain exist, including complex, quaternion, and octonion, among others. They are identified by the number of imaginary units they employ and by the properties of their vector multiplication.
The quaternion domain is one of the most popular for neural networks thanks to the Hamilton product properties. This domain has its foundations in the quaternion number $q = q_0 + q_1 \ii + q_2 \ij + q_3 \ik$, in which $q_c, \; c \in \{0,1,2,3\}$ are real coefficients and $\ii, \ij, \ik$ the imaginary untis. A quaternion with its real part $q_0$ equal to $0$ is named \textit{pure quaternion}. The imaginary units comply with the property $\ii^2 = \ij^2 = \ik^2 = -1$ and with the non-commutative products $\ii \ij = - \ij \ii ; \; \ij \ik = - \ik \ij ; \; \ik \ii = - \ii \ik$. Due to the non-commutativity of vector multiplication, the Hamilton product has been introduced to properly model the multiplication between two quaternions.

\subsection{Real and Quaternion-Valued Convolutional Layers}


A generic convolutional layer can be described by

\begin{equation}
    \y = \text{Conv}(\x) = \W * \x + \mathbf{b},
\label{eq:conv}
\end{equation}
where the input $\x \in \bR^{t \times s}$ is convolved ($*$) with the filters tensor $\W \in \bR^{s \times d \times k \times k}$ to produce the output $\y \in \bR^{d \times t}$, where $s$ is the input channels dimension, $d$ the output one, $k$ is the filter size, and $t$ is the input and output dimension. The bias term $\mathbf{b}$ does not heavily influence the number of parameters, thus the degrees of freedom for this operation are essentially $\mathcal{O}(sdk^2)$.

Quaternion convolutional layers, instead, build the weight tensor $\W \in \bR^{s \times d \times k \times k}$ by following the Hamilton product rule and organize filters according to it:

\begin{equation}
{\bf{W}} * {\bf{x}} = \left[ {\begin{array}{*{20}c}
   \hfill {{\bf{W}}_0 } & \hfill { - {\bf{W}}_1 } & \hfill { - {\bf{W}}_2 } & \hfill { - {\bf{W}}_3 } \\
   \hfill {{\bf{W}}_1 } & \hfill {{\bf{W}}_0 } & \hfill { - {\bf{W}}_3 } & \hfill {{\bf{W}}_2 } \\
   \hfill {{\bf{W}}_2 } & \hfill {{\bf{W}}_3 } & \hfill {{\bf{W}}_0 } & \hfill { - {\bf{W}}_1 } \\
   \hfill {{\bf{W}}_3 } & \hfill { - {\bf{W}}_2 } & \hfill {{\bf{W}}_1 } & \hfill {{\bf{W}}_0 } \\
\end{array}} \right] * \left[ {\begin{array}{*{20}c}
   {{\bf{x}}_0 } \hfill  \\
   {{\bf{x}}_1 } \hfill  \\
   {{\bf{x}}_2 } \hfill  \\
   {{\bf{x}}_3 } \hfill  \\
\end{array}} \right]
\label{eq:qprod}
\end{equation}


where $\W_0, \W_1, \W_2, \W_3 \in \bR^{\frac{s}{4} \times \frac{d}{4} \times k \times k}$ are the real coefficients of the quaternion weight matrix $\W = \W_0 + \W_1 \ii + \W_2 \ij + \W_3 \ik$  and $\x_0, \x_1, \x_2, \x_3$ are the coefficients of the quaternion input $\x$ with the same structure.

As done for real-valued layers, the bias can be ignored and the degree of freedom computations of the quaternion convolutional layer can be approximated to $\mathcal{O}(sdk^2/4)$. The lower number of parameters with respect to the real-valued operation is due to the reuse of filters performed by the Hamilton product in Eq.~\ref{eq:qprod}. Also, sharing the parameter submatrices forces to consider and exploit the correlation between the input components \cite{ParcolletAIR2019, Tay2019QTRansformer, GaudetIJCNN2018}.

\section{Parameterizing Hypercomplex Convolutions}
\label{sec:phc}

\begin{figure*}[t]
    \begin{center}
        \includegraphics[width=0.9\textwidth]{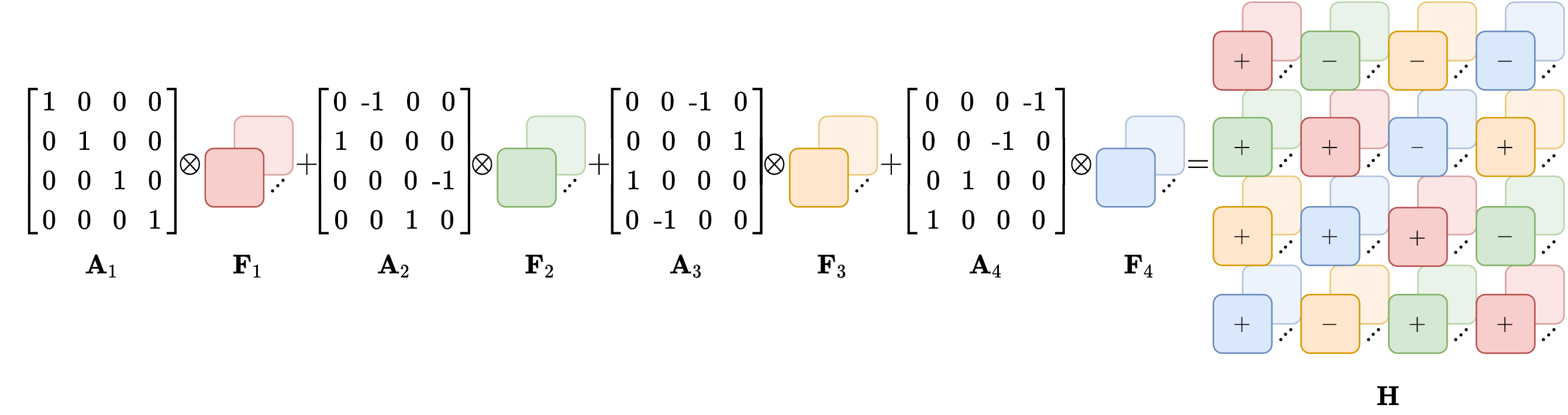}
    \end{center}
    \caption{The quaternion convolution rule can be expressed as sum of Kronecker products between the matrices $\mathbf{A}_i$ that subsume the algebra rules and the matrices $\mathbf{F}_i$ that contain the convolution filters, with $i=1,2,3,4$. In this example, the parameters of $\mathbf{A}_i$ are fixed for visualization purposes, but in PHC layers they are learnable parameters.}
    \label{fig:ham_prod}
\end{figure*}

In the following, we delineate the formulation for the proposed parameterized hypercomplex convolutional (PHC) layer. We also show that this approach is capable of learning the Hamilton product rule when two quaternions are convolved.

\subsection{Parameterized Hypercomplex Convolutional Layers}
\label{subsec:phc_layers}

The PHC layer is based on the construction, by sum of Kronecker products, of the weight tensor $\mathbf{H}$ which encapsulates and organizes the filters of the convolution. The proposed method is formally defined as:
\begin{equation}
    \y = \text{PHC}(\x) = \mathbf{H}*\x + \mathbf{b},
\end{equation}

\noindent whereby, $\mathbf{H} \in \bR^{s \times d \times k \times k}$  is built by sum of Kronecker products between two learnable groups of matrices. Here, $s$ is the input dimensionality to the layer, $d$ is the output one, and $k$ is the filter size.
More concretely,
\begin{equation}
    \mathbf{H} = \sum_{i=1}^n \mathbf{A}_i \otimes \mathbf{F}_i,
\end{equation}
\noindent in which $\mathbf{A}_i \in \bR^{n \times n}$ with $i=1, ..., n$ are the matrices that describe the algebra rules and $\mathbf{F}_i \in \bR^{\frac{s}{n} \times \frac{d}{n} \times k \times k}$ represents the $i$-th batch of filters that are arranged by following the algebra rules to compose the final weight matrix. It is worth noting that $\frac{s}{n} \times \frac{d}{n} \times k \times k$ holds for squared kernels, while $\frac{s}{n} \times \frac{d}{n} \times k$ should be considered instead for 1D kernels. The core element of this approach is the Kronecker product \cite{KroneckerBook}, which is a generalization of the vector outer product that can be parameterized by $n$. The hyperparameter $n$ can be set by the user who wants to operate in a pre-defined real or hypercomplex domain (e.g., by setting $n=2$ the PHC layer is defined in the complex domain, or in the quaternion one if $n$ is set equal to $4$, as Figure \ref{fig:ham_prod} illustrates), or tuned to obtain the best performance from the model. The matrices $\mathbf{A}_i$ and $\mathbf{F}_i$ are learnt during training and their values are reused to build the definitive tensor $\mathbf{H}$. 

The degree of freedom of $\mathbf{A}_i$ and $\mathbf{F}_i$ are $n^3$ and $sdk^2/n$, respectively. Usually, real world applications employ a large number of filters in layers ($s, d = 256, 512, ...)$ and small values for $k$. Therefore, frequently $sdk^2 \gg n^3$ holds. Thus, the degrees of freedom for the PHC weight matrix can be approximated to $\mathcal{O}(sdk^2/n)$. Hence, the PHC layer reduces the number of parameters by $1/n$ with respect to a standard convolutional layer in real world problems.

\begin{figure}[t]
    \begin{center}
        \includegraphics[width=\linewidth]{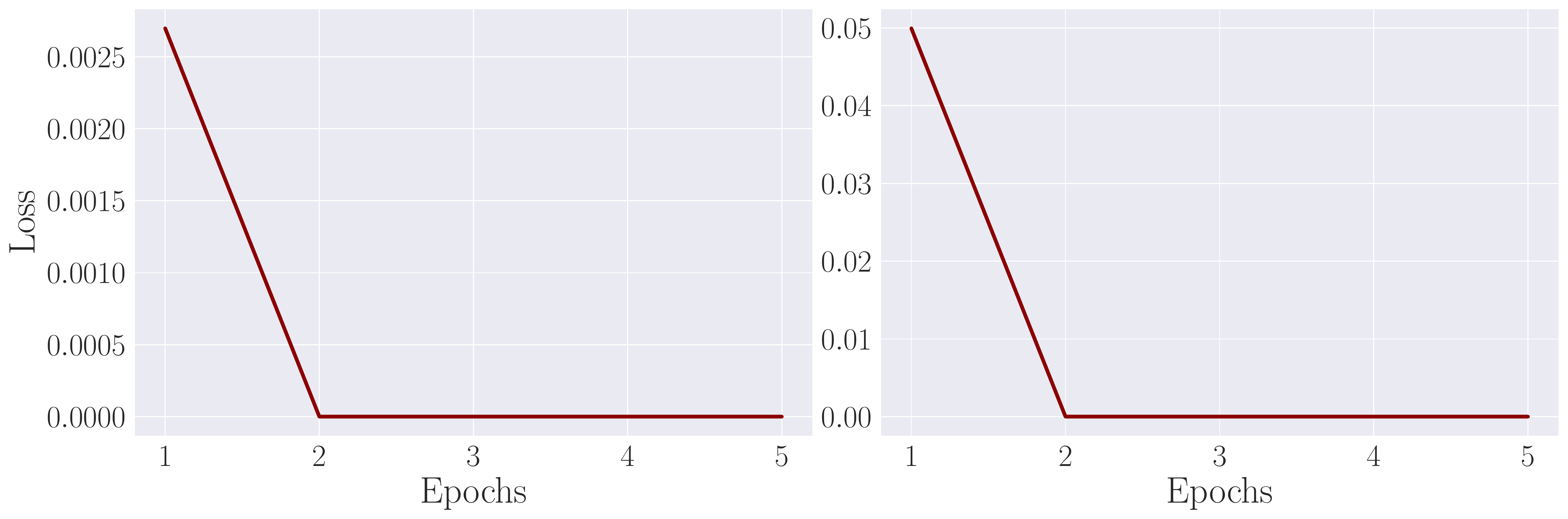}
    \end{center}
    \caption{Loss plots for toy examples. The PHC layer is able to learn the matrix $\mathbf{A}$ describing the convolution rule for pure (left) and full quaternions (right).}
    \label{fig:toy}
\end{figure}

Moreover, when processing multidimensional data with correlated channels, such as color images, rather than mulichannel audio or multisensor signals, PHC layers bring benefits due to the weight sharing among different channels. This allows capturing latent intra-channels relations that standard convolutional networks ignore because of the rigid structure of the weights \cite{GrassucciQGAN2021, ParcolletICASSP2019a}. The PHC layer is able to subsume hypercomplex convolution rules and the desired domain is specified by the hyperparameter $n$.
Interestingly, by setting $n=1$ a real-valued convolutional layer can be represented too. Indeed, standard real layers do not involve parameter sharing, therefore the algebra rules are solely described by the single $\mathbf{A} \in \bR^{1\times1}$ and the complete set of filters are included in $\mathbf{F}^{s \times d \times k \times k}$.

Therefore, the PHC layer fills the gaps left by pre-existing hypercomplex algebras in Fig.~\ref{fig:hprod} and subsumes the missing algebra rules directly from data, i.e., the dashed grey lines in Fig.~\ref{fig:hprod}. Thus, a neural model equipped with PHC layers can grasp the filter organization also for $n=3,5,6,7$ and so on.
Moreover, any convolutional model can be endowed with our approach, since PHC layers easily replace standard convolution / transposed convolution operations and the hyperparameter $n$ gives high flexibility to adapt the layer to any kind of input, such as color images, multichannel audio or multisensor signals. 
\subsection{Learning Tests on Toy Examples}

We test the receptive ability of the PHC layer in two toy problems building an artificial dataset. We highly encourage the reader to take a look at the section \texttt{tutorials} of the GitHub repository \url{https://github.com/eleGAN23/HyperNets} for more insights and results on toy examples, including the learned matrices $\mathbf{A}_i$. The first task aims at learning the right matrix $\mathbf{A}$  to build a quaternion convolutional layer which properly follows the Hamilton rule in Eq.~\ref{eq:qprod}. That is, we set $n=4$ and the objective is to learn the four matrices $\mathbf{A}_i$ as they are in the quaternion product in Fig.~\ref{fig:ham_prod}. We build the dataset by performing a convolution with a matrix of filters $\W \in \bH$, which are arranged following the regulation in Eq.~\ref{eq:qprod}, and a quaternion $\x \in \bH$ in input. The target is still a quaternion, named $\mathbf{y} \in \bH$. As shown in Fig. \ref{fig:toy} (right), the MSE loss of the PHC layer converges very fast, meaning that the layer properly learns the matrix $\mathbf{A}$ and the Hamilton convolution. 

The second toy example is a modification of the previous dataset target. Here, we want to learn the matrix $\mathbf{A}$ which describes the convolution among two pure quaternions. Therefore, when setting $n=4$, the matrix $\mathbf{A}_1$ of a pure quaternion should be complete null. Pure quaternions may be, as an example, an input RGB image and the weights of a hypercomplex convolutional layer since the first channel of RGB images is zero. Figure \ref{fig:toy} (left) displays the convergence of the PHC layer loss during training, proving that the proposed method is able of subsuming hypercomplex convolutional rules when dealing with pure quaternions too.

\begin{strip}
\begin{gather}
\label{eq:visualphc}
\begin{array}{c}
 \mathop {\left[ A \right]}\limits_{\left( {1 \times 1} \right)}  \otimes \mathop {\left[ {\begin{array}{*{20}c}
   {}  \\
   {}  \\
   {}  \\
   {}  \\
   {}  \\
   {}  \\
   {}  \\
\end{array}\begin{array}{*{20}c}
   {}  \\
   {}  \\
   {}  \\
   {}  \\
   {}  \\
   {}  \\
   {}  \\
\end{array}\begin{array}{*{20}c}
   {}  \\
   {}  \\
   {}  \\
   {}  \\
   {}  \\
   {}  \\
   {}  \\
\end{array}{\bf{F}}\begin{array}{*{20}c}
   {}  \\
   {}  \\
   {}  \\
   {}  \\
   {}  \\
   {}  \\
   {}  \\
\end{array}\begin{array}{*{20}c}
   {}  \\
   {}  \\
   {}  \\
   {}  \\
   {}  \\
   {}  \\
   {}  \\
\end{array}\begin{array}{*{20}c}
   {}  \\
   {}  \\
   {}  \\
   {}  \\
   {}  \\
   {}  \\
   {}  \\
\end{array}} \right]}\limits_{\left( {s \times d \times k \times k} \right)}  = \mathop {\left[ {\begin{array}{*{20}c}
   {}  \\
   {}  \\
   {}  \\
   {}  \\
   {}  \\
   {}  \\
   {}  \\
\end{array}\begin{array}{*{20}c}
   {}  \\
   {}  \\
   {}  \\
   {}  \\
   {}  \\
   {}  \\
   {}  \\
\end{array}\begin{array}{*{20}c}
   {}  \\
   {}  \\
   {}  \\
   {}  \\
   {}  \\
   {}  \\
   {}  \\
\end{array}{\bf{H}}\begin{array}{*{20}c}
   {}  \\
   {}  \\
   {}  \\
   {}  \\
   {}  \\
   {}  \\
   {}  \\
\end{array}\begin{array}{*{20}c}
   {}  \\
   {}  \\
   {}  \\
   {}  \\
   {}  \\
   {}  \\
   {}  \\
\end{array}\begin{array}{*{20}c}
   {}  \\
   {}  \\
   {}  \\
   {}  \\
   {}  \\
   {}  \\
   {}  \\
\end{array}} \right]}\limits_{\left( {s \times d \times k \times k} \right)}  \\ 
 \mathop {\left[ {{\bf{A}}_1 } \right]}\limits_{\left( {2 \times 2} \right)}  \otimes \mathop {\left[ {\begin{array}{*{20}c}
   {}  \\
   {}  \\
   {}  \\
\end{array}{\bf{F}}_1 \begin{array}{*{20}c}
   {}  \\
   {}  \\
   {}  \\
\end{array}} \right]}\limits_{\left( {\frac{s}{2} \times \frac{d}{2} \times k \times k} \right)}  + \mathop {\left[ {{\bf{A}}_2 } \right]}\limits_{\left( {2 \times 2} \right)}  \otimes \mathop {\left[ {\begin{array}{*{20}c}
   {}  \\
   {}  \\
   {}  \\
\end{array}{\bf{F}}_2 \begin{array}{*{20}c}
   {}  \\
   {}  \\
   {}  \\
\end{array}} \right]}\limits_{\left( {\frac{s}{2} \times \frac{d}{2} \times k \times k} \right)}  = \mathop {\left[ {\begin{array}{*{20}c}
   {}  \\
   {}  \\
   {}  \\
   {}  \\
   {}  \\
   {}  \\
   {}  \\
\end{array}\begin{array}{*{20}c}
   {}  \\
   {}  \\
   {}  \\
   {}  \\
   {}  \\
   {}  \\
   {}  \\
\end{array}\begin{array}{*{20}c}
   {}  \\
   {}  \\
   {}  \\
   {}  \\
   {}  \\
   {}  \\
   {}  \\
\end{array}{\bf{H}}\begin{array}{*{20}c}
   {}  \\
   {}  \\
   {}  \\
   {}  \\
   {}  \\
   {}  \\
   {}  \\
\end{array}\begin{array}{*{20}c}
   {}  \\
   {}  \\
   {}  \\
   {}  \\
   {}  \\
   {}  \\
   {}  \\
\end{array}\begin{array}{*{20}c}
   {}  \\
   {}  \\
   {}  \\
   {}  \\
   {}  \\
   {}  \\
   {}  \\
\end{array}} \right]}\limits_{\left( {s \times d \times k \times k} \right)}  \\ 
 \begin{array}{*{20}c}
    \vdots   \\
    \vdots   \\
\end{array} \\ 
 \mathop {\left[ {{\bf{A}}_1 } \right]}\limits_{\left( {n \times n} \right)}  \otimes \mathop {\left[ {\begin{array}{*{20}c}
   {}  \\
\end{array}{\bf{F}}_1 \begin{array}{*{20}c}
   {}  \\
\end{array}} \right]}\limits_{\left( {\frac{s}{n} \times \frac{d}{n} \times k \times k} \right)}  + \mathop {\left[ {{\bf{A}}_2 } \right]}\limits_{\left( {n \times n} \right)}  \otimes \mathop {\left[ {\begin{array}{*{20}c}
   {}  \\
\end{array}{\bf{F}}_2 \begin{array}{*{20}c}
   {}  \\
\end{array}} \right]}\limits_{\left( {\frac{s}{n} \times \frac{d}{n} \times k \times k} \right)}  +  \ldots  + \mathop {\left[ {{\bf{A}}_n } \right]}\limits_{\left( {n \times n} \right)}  \otimes \mathop {\left[ {\begin{array}{*{20}c}
   {}  \\
\end{array}{\bf{F}}_n \begin{array}{*{20}c}
   {}  \\
\end{array}} \right]}\limits_{\left( {\frac{s}{n} \times \frac{d}{n} \times k \times k} \right)}  = \mathop {\left[ {\begin{array}{*{20}c}
   {}  \\
   {}  \\
   {}  \\
   {}  \\
   {}  \\
   {}  \\
   {}  \\
\end{array}\begin{array}{*{20}c}
   {}  \\
   {}  \\
   {}  \\
   {}  \\
   {}  \\
   {}  \\
   {}  \\
\end{array}\begin{array}{*{20}c}
   {}  \\
   {}  \\
   {}  \\
   {}  \\
   {}  \\
   {}  \\
   {}  \\
\end{array}{\bf{H}}\begin{array}{*{20}c}
   {}  \\
   {}  \\
   {}  \\
   {}  \\
   {}  \\
   {}  \\
   {}  \\
\end{array}\begin{array}{*{20}c}
   {}  \\
   {}  \\
   {}  \\
   {}  \\
   {}  \\
   {}  \\
   {}  \\
\end{array}\begin{array}{*{20}c}
   {}  \\
   {}  \\
   {}  \\
   {}  \\
   {}  \\
   {}  \\
   {}  \\
\end{array}} \right]}\limits_{\left( {s \times d \times k \times k} \right)}.  \\ 
 \end{array}
\end{gather}
\end{strip}

\subsection{Demystifying Parameterized Hypercomplex Convolutional Layers}
\label{subsec:dem_phc}

We provide a formal explanation of the PHC layer to better understand the Kronecker product and how it organizes convolution filters to reduce the overall number of parameters to $1/n$. In Eq.~\ref{eq:visualphc}, we show how the PHC layer generalizes from $1$D to $n$D domains. When subsuming real-valued convolutions in the first line of Eq.~\ref{eq:visualphc}, the Kronecker product is performed between a scalar $A$ and the filter matrix $\mathbf{F}$, whose dimension is the same as the final weight matrix $\mathbf{H}$, which is $s \times d \times k \times k$.

Considering the complex case with $n=2$ in the second line of Eq.~\ref{eq:visualphc}, the algebra is defined in $\mathbf{A}_1$ and $\mathbf{A}_2$ while the filters are contained in $\mathbf{F}_1$ and $\mathbf{F}_2$, each of dimension $1/2$ the final matrix $\mathbf{H}$.
Therefore, while the size of the weight matrix $\mathbf{H}$ remains unchanged, the parameter size is approximately $1/2$ the real one. In the last line of Eq.~\ref{eq:visualphc}, we can see the generalization of this process, in which the size of matrices $\mathbf{F}_i$, $i=1, ..., n$ is reduced proportionally to $n$. It is worth noting that, while the parameter size is reduced with growing values of $n$, the dimension of $\mathbf{H}$ remains the same.


\section{Parameterized Hypercomplex Neural Networks for Color Images}
\label{sec:phc_rgb}

In this section, we describe how PHNNs can be applied for processing color images in hypercomplex domains without needing any additional information to the input and we propose examples of parameterized hypercomplex versions of common computer vision models such as VGGs and ResNets. In order to be consistent with literature, we perform each experiment with a real-valued baseline model, then we compare it with its complex and quaternion counterparts and with the proposed PHNN. Furthermore, we assess the malleability of the proposed approach testing different values of the hyperparameter $n$, therefore defining parameterized hypercomplex models in multiple domains.

\subsection{Process Color Images with PHC Layers}

Different encodes exist to process color images, however, the most common computer vision datasets are comprised of three-channel images in $\bR^3$. In the quaternion domain, RGB images are enclosed into a quaternion and processed as single elements \cite{ParcolletAIR2019}. The encapsulation is performed by considering the RGB channels as the real coefficients of the imaginary units and by padding a zeros channel as the first real component of the quaternion.

Here, we propose to leverage the high malleability of PHC layers to deal with RGB images in hypercomplex domains without embedding useless information to the input. Indeed, the PHC can directly operate in $\bR^3$ by easily setting $n=3$ and process RGB images in their natural domain while exploiting hypercomplex network properties such as parameters sharing. Indeed, the great flexibility of PHC layers allows the user to choose whether processing images in $\bR^4$ or $\bR^3$. On one hand, by setting $n=4$, the zeros channel is added to the input even so the layer saves the $75\%$ of free parameters. On the other hand, by choosing $n=3$ the network does not handle any useless information, notwithstanding, it reduces the number of parameters by solely $66\%$. This is a trade-off which may depend on the application or on the hardware the user needs. Furthermore, the domain on which processing images can be tuned by letting the performance of the network indicates the best choice for $n$.

\subsection{Parameterized Hypercomplex VGGs}
A family of popular methods for image processing is based on the VGG networks \cite{VGG2015} that stack several convolutional layers and a closing fully connected classifier. To completely define models in the desired hypercomplex domain, we propose to endow the network with PHC layers as convolution components and with Parameterized Hypercomplex Multiplication (PHM) layers \cite{Zhang2021PHM} as linear classifier. The backbone of our PHVGG is then
\begin{equation}
    \begin{split}
    \mathbf{h}_t &= \text{ReLU} \left( \text{PHC}_t \left( \mathbf{h}_{t-1} \right) \right) \qquad t=1,...,j \\
    \y &= \text{ReLU} \left( \text{PHM}(\mathbf{h}_j) \right).
    \end{split}
\end{equation}


\subsection{Parameterized Hypercomplex ResNets}
In recent literature, a copious set of high performance in image classification is obtained with models having a residual structure. ResNets \cite{Resnet2016} pile up manifold residual blocks composed of convolutional layers and identity mappings. A generic PHResNet residual block is defined by
\begin{equation}
    \y = \mathcal{F}(\x, \{ \mathbf{H}_j \}) + \x,
\end{equation}

whereby $\mathbf{H}_j$ are the PHC weights of layer $j = 1, 2$ in the block, and $\mathcal{F}$ is
\begin{equation}
    \mathcal{F}(\x, \{ \mathbf{H}_j \}) = \text{PHC} \left( \text{ReLU} \left( \text{PHC}(\x ) \right) \right),
\end{equation}

\noindent in which we omit batch normalization to simplify notation.
The backward phase of a PHNNs reduces to a backpropagation similar to the quaternion neural networks one, which has been already developed in \cite{NittaQBack1995, ParcolletAIR2019, ParcolletICLR2019}.

\section{Parameterized Hypercomplex Neural Networks for Multichannel Signals}
\label{sec:phc_audio}

In the following, we expound how PHNNs can be employed to deal with multichannel audio signals and we introduce, as an example, the parameterized hypercomplex Sound Event Detection networks (PHSEDnets).

\subsection{Process multichannel audio with PHC layers}
A first-order Ambisonics (FOA) signal is composed of $4$ microphone capsules, whose magnitude representations can be enclosed in a quaternion \cite{ComminielloICASSP2019a, RicciardiMLSP2020}. However, the quaternion algebra may be restrictive if more than one microphone is employed for registration or whether the phase information has to be included too. Indeed, quaternion neural networks badly fit with multidimensional input with more than $4$ channels \cite{Grassucci2022DualQ}.

Conversely, the proposed method can be easily adapted to deal with these additional dimensions by handily setting the hyperparameter $n$ and thus completely leveraging each information in the $n$-dimensional input.

\subsection{Parameterized Hypercomplex SEDnets}

Sound Event Detection networks (SEDnets) \cite{Adavanne2019SoundEL} are comprised of a core convolutional component which extracts features from the input spectrogram. The information is then passed to a gated recurrent unit (GRU) module and to a stack of fully connected (FC) layers with a closing sigmoid $\sigma$ which outputs the probability the sound is in the audio frame. Formally, the PHSEDnet is described by
\begin{equation}
    \begin{split}
    \mathbf{h}_t &= \text{PHC}_t(\mathbf{h}_{t-1}) \qquad t=1,...,j\\
    \y &= \sigma \left( \text{FC} \left( \text{GRU} \left( \mathbf{h}_j \right) \right) \right).
    \end{split}
\end{equation}
After the GRU model, We employ standard fully connected layers, that can be also implemented as PHM layers with $n=1$, since the so processed signal loses its multidimensional original structure.


\section{Experimental Evaluation on Image Classification}
\label{sec:img_class}

\begin{figure}[t]
    \centering
    \includegraphics[width=\linewidth]{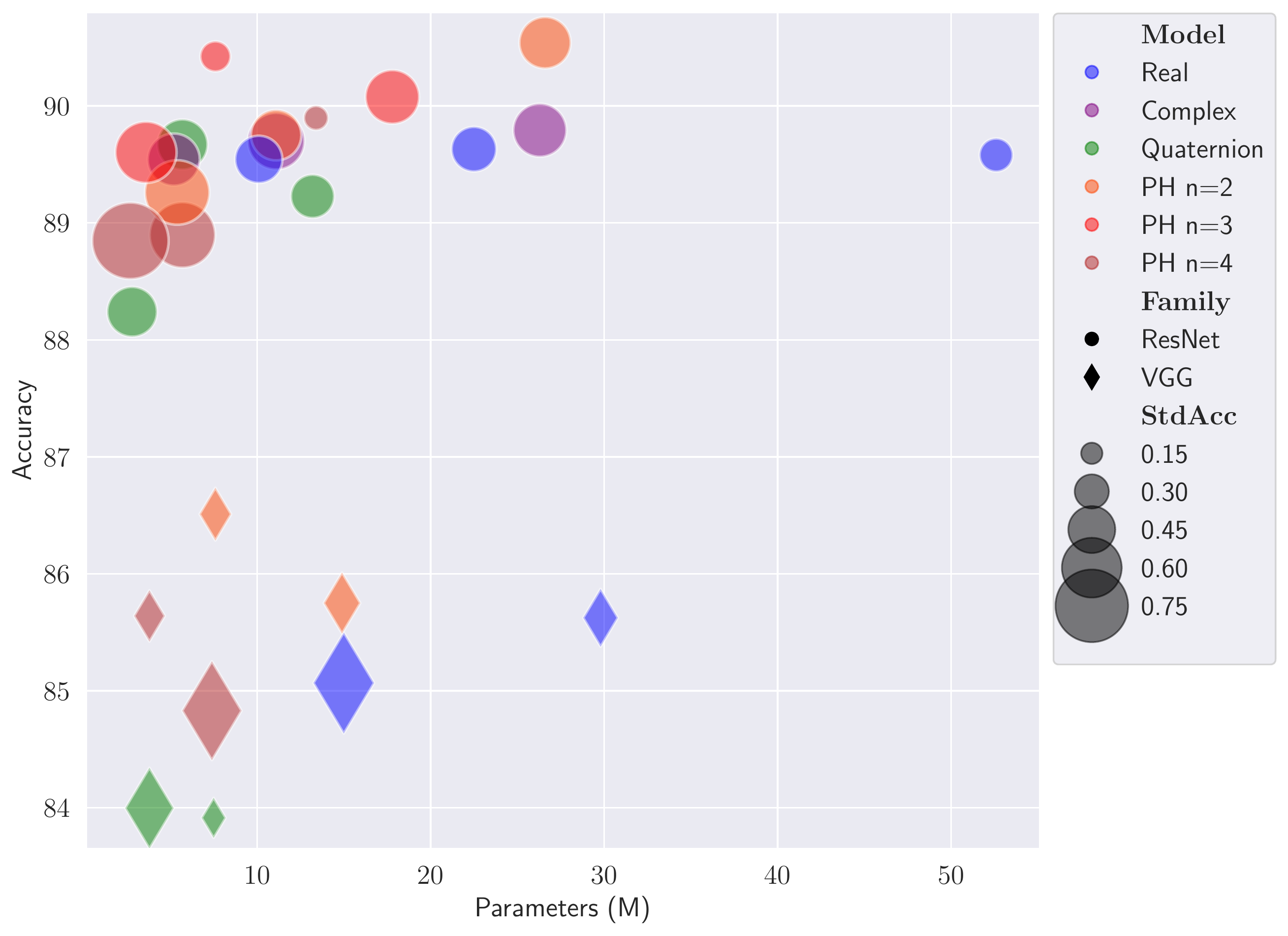}
    \caption{CIFAR10 accuracy against number of network parameters for VGG and ResNet models. The larger is the point, the higher is the standard deviation over the runs. PHC-based models obtain better accuracies in both the families while far reducing the number of parameters. We do not display Complex VGGs as their accuracy is very low with respect to other models.}
    \label{fig:bubble}
\end{figure}

To begin with, we test the PHC layer on RGB images and we show how, exploiting the correlations among channels, the proposed method saves parameters while ensuring high performance. We perform each experiment with a real-valued baseline model and then we compare it with its complex and quaternion counterparts and with the proposed PHNNs. Furthermore, we assess the malleability of the proposed approach testing different values of the hyperparameter $n$, therefore defining parameterized hypercomplex models in multiple domains.

\subsection{Experimental Setup}
We perform the image classification task with five baseline models. We consider ResNet18, ResNet50 and ResNet152 from the ResNet family and VGG16 and VGG19 from the VGG one. Each hyperparameter is set according to the original papers \cite{Resnet2016, VGG2015}. We investigate the performance in four different color images datasets at different scales. We employ SVHN, CIFAR10, CIFAR100, and ImageNet and any kind of data augmentation is applied to these datasets in order to guarantee a fair comparison.

We modify the number of filters for ResNets in order to be divisible by $3$ and thus having the possibility of testing a configuration with $n=3$. The modified versions of the ResNets are built with an initial convolutional layer of $60$ filters. Then, the subsequent blocks have $60, 120, 240, 516$ filters. The number of layers in the blocks depends on the ResNet chosen, whether 18, 50 or 152. Instead, VGG19 convolution component comprise two $24$, two $72$, four $216$, and eight $648$ filter layers, with batch normalization. The classifier is composed of three fully connected layers of $648$, $516$ and $10$, $100$ or $1000$ depending on the number of classes in the dataset. The rest of the hyperparameters are set as suggested in the original papers.
The batch size is fixed to $128$ and training is performed via SGD optimizer with momentum equal to $0.9$, weight decay $5e^{-4}$ and a cosine annealing scheduler. For ResNets, the initial learning rate is set to $0.1$. For VGG is equal to $0.01$. Models on CIFAR10 and CIFAR100 are trained for $200$ epochs whereas on SVHN networks run for $50$ epochs. For the ImageNet dataset, we follow the recipes in \cite{wightman2021resnet}, so we resize the images for training at $160\times160$ while keeping the standard size of $224\times224$ for validation and test. We employ a step learning rate decay every $30$ epochs with $\gamma = 0.1$, the SGD optimizer and an initial learning rate of $0.1$ with weight decay $0.0001$. The training is performed for $300$k iterations with a batch size of $256$ employing four Tesla V100 GPUs.

\subsection{Experimental Results}
\begin{figure}[t]
    \centering
    \includegraphics[width=\linewidth]{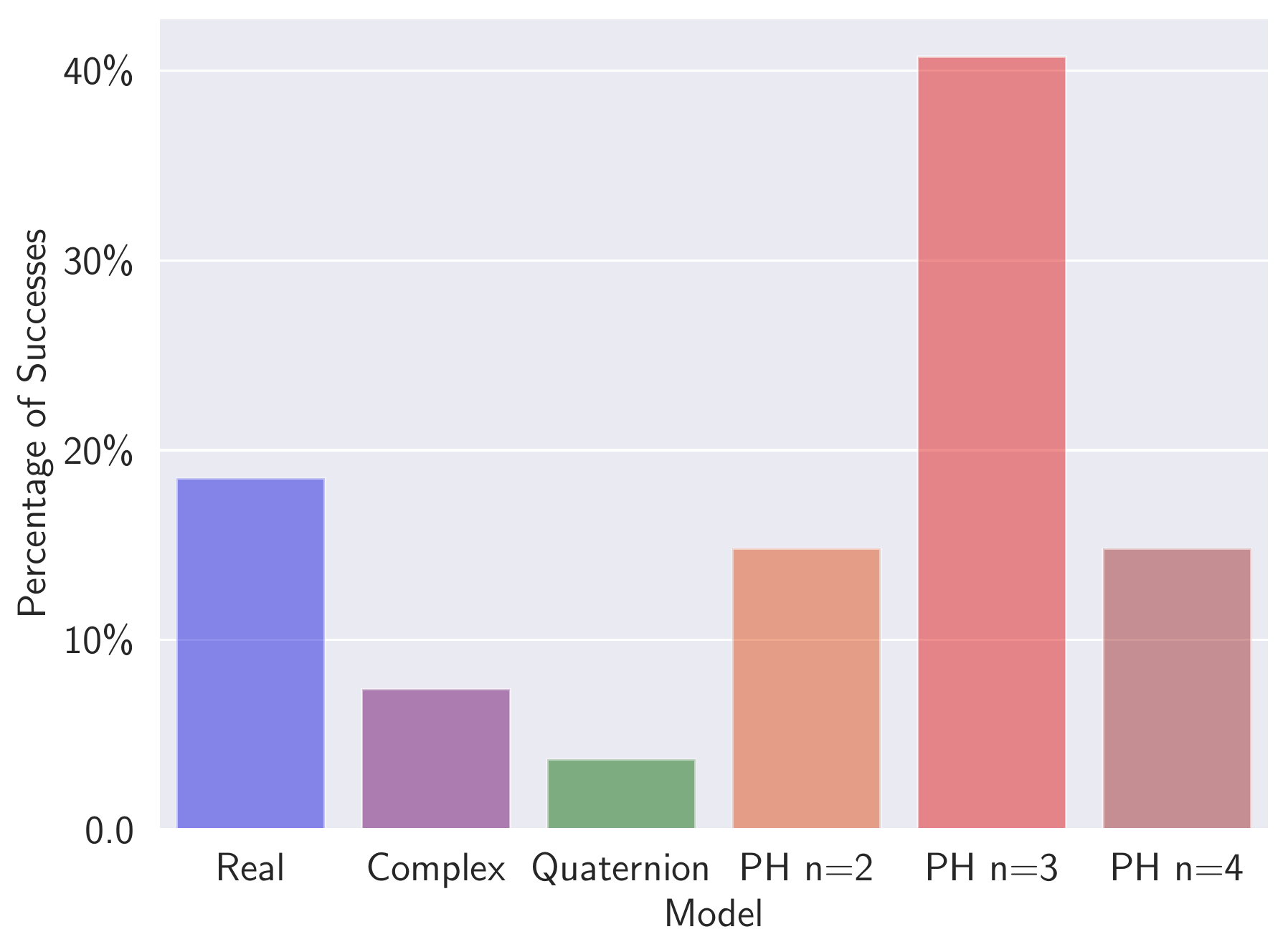}
    \caption{Bar plot of number of successes achieves by the models in Table~\ref{tab:img_class_full} in each of the runs. The PHC-based models with $n=3$ (red bar) far exceeds other configurations being the more performing choice for RGB image classification task.}
    \label{fig:barplot}
\end{figure}

\begin{table*}[t]
\caption{Image classification results for VGG. The accuracy mean and standard deviation over three runs with different seeds is reported. Training (T) time and inference (I) time required on CIFAR10. For training time we report, in seconds per 100 iterations, the mean and the standard deviation over the iterations in one epoch, while the inference time is the time required to decode the test set. The PHNN with $n=4$ outperforms the quaternion counterpart both in terms of accuracy and time. The PHVGG with $n=2$ far exceeds the real-valued baseline in the considered datasets, while both the PHVGG19 versions with $n=2,4$ are more efficient than the real and quaternion-valued baselines at inference time. $p$-value under the T-test $0.0002$.}
\label{tab:img_class}
\begin{center}
\begin{tabular}{llcccc}
\multicolumn{1}{c}{\bf Model} &\multicolumn{1}{c}{\bf Params} &\multicolumn{1}{c}{\bf SVHN} &\multicolumn{1}{c}{\bf CIFAR10} &\multicolumn{1}{c}{\bf Time (T)} &\multicolumn{1}{c}{\bf Time (I)}\\
\hline \\
VGG16         & 15M          & 94.364 $\pm$ 0.394 & 85.067 $\pm$ 0.765 & \textbf{2.2 $\pm$ 0.02} & \textbf{1.2}\\
Complex VGG16 & 7.6M (-50\%) & 93.555 $\pm$ 0.392 & 76.927 $\pm$ 0.511 & 5.2 $\pm$ 0.02 & 1.5\\
Quaternion VGG16 & 3.8M (-75\%) & 93.887 $\pm$ 0.292 & 83.997 $\pm$ 0.493 & 5.2 $\pm$ 0.02 & 2.2\\
PHVGG16 $n=2$ & 7.6M (-50\%) & \textbf{94.831 $\pm$ 0.257} & \textbf{86.510 $\pm$ 0.216} & \underline{3.2 $\pm$ 0.02} & \underline{1.4}\\
PHVGG16 $n=4$ & 3.8M (-75\%) & \underline{94.639 $\pm$ 0.121}  & \underline{85.640 $\pm$ 0.205} &  \underline{3.2 $\pm$ 0.02} & \underline{1.4}\\
\hline
VGG19                & 29.8M &   94.140 $\pm$ 0.129    &     \underline{85.624 $\pm$ 0.257}    &  \textbf{3.2 $\pm$ 0.02} & 16.0   \\
Complex VGG19 & 14.8M (-50\%) & 90.469 $\pm$ 0.222 & 76.979 $\pm$ 0.345 & 5.2 $\pm$ 0.02 & 16.2\\
Quaternion VGG19     & 7.5M (-75\%)  &   93.983 $\pm$ 0.190   &    83.914 $\pm$ 0.129     &   6.2 $\pm$ 0.02 & 16.3  \\
PHVGG19 $n=2$      & 14.9M (-50\%) &  \textbf{94.553 $\pm$ 0.229}    &    \textbf{85.750 $\pm$ 0.286}     &   \underline{4.0 $\pm$ 0.02} & \textbf{15.4}  \\
PHVGG19 $n=4$      & 7.4M (-75\%)  &  \underline{94.169 $\pm$ 0.296}    &    84.830 $\pm$ 0.733     &  4.2 $\pm$ 0.02 & \underline{15.5}   \\
\end{tabular}
\end{center}
\end{table*}

\begin{table*}[t]
\caption{Image Classification results with ResNet models. Each experiment is run three times with different seeds and mean with standard deviation is reported. The proposed models far exceed real-valued and quaternion baselines almost in each experiment we conduct. Interestingly, the PHNN outperform the real-valued counterpart by $4\%$ points in the largest-scale experiment on CIFAR100. The time is similar to the claims in Table \ref{tab:img_class} so we do not add here to avoid redundancy.}
\label{tab:img_class_full}
\begin{center}
\begin{tabular}{lllccccc}
\multicolumn{1}{c}{\bf Model} &\multicolumn{1}{c}{\bf Params} &\multicolumn{1}{c}{\bf Storage Memory} &\multicolumn{1}{c}{\bf SVHN} &\multicolumn{1}{c}{\bf CIFAR10} &\multicolumn{1}{c}{\bf CIFAR100} \\
\hline \\
ResNet18             & 10.1M & 39MB &   93.992 $\pm$ 1.317   &    \underline{89.543 $\pm$ 0.340}     &     \underline{62.634 $\pm$ 0.600}    \\
Complex ResNet18  & 5.2M (-50\%) & 20MB (-50\%) &   89.902 $\pm$ 0.322   &    89.541 $\pm$ 0.412     &    60.417 $\pm$ 0.811    \\
Quaternion ResNet18  & 2.8M (-75\%) & 10MB (-75\%) &   93.661 $\pm$ 0.413   &    88.240 $\pm$ 0.377     &    59.850 $\pm$ 0.607    \\
PHResNet18 $n=2$   & 5.4M (-50\%) & 20MB (-50\%) &   \textbf{94.359 $\pm$ 0.187}   &    89.260 $\pm$ 0.625     &    60.320 $\pm$ 2.249  \\
PHResNet18 $n=3$   & 3.6M (-66\%) & 13MB (-66\%) &  \underline{94.303 $\pm$ 1.234}   &    \textbf{89.603 $\pm$ 0.563}     &     \textbf{62.660 $\pm$ 1.067}    \\
PHResNet18 $n=4$   & 2.7M (-75\%) & 10MB (-75\%) &  94.234 $\pm$ 0.161    &    88.847 $\pm$ 0.874     &    61.780 $\pm$ 0.689    \\
\hline
ResNet50             & 22.5M & 86MB &   \underline{94.546 $\pm$ 0.269}   &    89.630 $\pm$ 0.305     &    65.514 $\pm$ 0.569      \\
Complex ResNet50  & 11.1M (-50\%) & 43MB (-50\%) &   89.004 $\pm$ 0.215   &    89.699 $\pm$ 0.485     &    65.104 $\pm$ 0.598    \\
Quaternion ResNet50  & 5.7M (-75\%) & 22MB (-75\%) &   93.685 $\pm$ 0.389  &    89.670 $\pm$ 0.383     &     63.760 $\pm$ 0.717     \\
PHResNet50 $n=2$   & 11.1M (-50\%) & 43MB (-50\%) &  93.849 $\pm$ 0.249   &    \underline{89.750 $\pm$ 0.386}     &    65.884 $\pm$ 0.333      \\
PHResNet50 $n=3$   & 7.6M (-66\%) & 29MB (-65\%) &   93.617 $\pm$ 0.497   &    \textbf{90.423 $\pm$ 0.145}     &    \textbf{66.497 $\pm$ 1.256}       \\
PHResNet50 $n=4$   & 5.7M (-75\%) & 23MB (-74\%) &  \textbf{94.558 $\pm$ 0.754}   &    88.897 $\pm$ 0.645     &     \underline{66.240 $\pm$ 1.165}     \\
\hline
ResNet152            & 52.6M & 201MB &  \textbf{94.625 $\pm$ 0.355}   &    89.580 $\pm$ 0.173     &     62.053 $\pm$ 0.385    \\
Complex ResNet152  & 26.3M (-50\%) & 101MB (-50\%) &   90.332 $\pm$ 0.129   &    89.792 $\pm$ 0.427     &    63.125 $\pm$ 0.681    \\
Quaternion ResNet152 & 13.2M (-75\%) & 51MB (-75\%) &  93.638 $\pm$ 0.098   &    89.227 $\pm$ 0.287     &   61.267 $\pm$ 0.784      \\
PHResNet152 $n=2$  & 26.6M (-50\%) & 103MB (-49\%) &  93.915 $\pm$ 0.512   &    \textbf{90.540 $\pm$ 0.401}     &      65.817 $\pm$ 0.327    \\
PHResNet152 $n=3$  & 17.8M (-66\%) & 70MB (-65\%) &  93.955 $\pm$ 0.152   &    \underline{90.077 $\pm$ 0.436}     &     \underline{66.347 $\pm$ 0.567}     \\
PHResNet152 $n=4$  & 13.4M (-75\%) & 53 MB (-74\%) &   \underline{94.290 $\pm$ 0.237}   &     89.897 $\pm$ 0.097    &     \textbf{66.437 $\pm$ 0.064}     \\
\end{tabular}
\end{center}
\end{table*}

We execute initial experiments with VGGs against Quaternion VGGs and two versions of PHVGGs with $n$ equal to $2$ and to $4$. Average and standard deviation accuracy over three runs are reported for SVHN and CIFAR10 datasets in Table \ref{tab:img_class}. We experiment also additional runs but any significant difference emerges as the randomness only affects the network initialization. Both the PHVGG16 and PHVGG19 versions clearly outperform real, complex and quaternion counterparts while being built with more than a half the number of parameters of the baseline. Additionally, PH-based models extraordinarily reduce the number of training and inference time (computed on an NVIDIA Tesla-V100) required with respect to the quaternion model which operates in a hypercomplex domain as well. Furthermore, when scaling up the experiment with VGG19, the proposed methods are more efficient at inference time with respect to the real-valued VGG19. Therefore, PHNNs can be easily adopted in applications with disk memory limitations, due to the reduction of parameters, and for fast inference problems thanks to the efficiency at testing time. Although the sum of Kronecker products in PHC layers requires additional computations, the increase is insignificant with respect to the FLOPs computated for the whole network, so the overall number of FLOPs is not heavily affected by our method and the count remains almost the same.

Our approach has high malleability, indeed, when dealing with color images, we can the domain in which operating thanks to the hyperparameter $n$. Therefore, we test PHNNs in the complex ($n=2$), quaternion ($n=4$) or $\bH^3$ ($n=3$) domain, where in the latter we do not concatenate any zero padding and process the RGB channels of the image in their natural domain.

Table \ref{tab:img_class_full} presents average and standard deviation accuracy over three runs with different seeds for ResNet-based models. We perform extensive experiments and the PH models with $n=4$ always outperform the quaternion counterpart gaining a higher accuracy and being more robust. This underlines the effectiveness of the PHC architectural flexibility over the predefined and rigid structure of quaternion layers. Furthermore, our method distinctly far exceeds the corresponding real-valued baselines across the experiments while saving from $50\%$ to $75\%$ parameters. Focusing on the latter result, the PHResNets with $n=3$ results to be the most suitable choice in many cases, proving the validity of processing RGB images in their natural domain leveraging hypercomplex algebra. However, performance with $n=3$ and $n=4$ are comparable, thus the choice of this hyperparameter may depend on the application or on the hardware employed. On one hand, $n=4$ may sometimes lead to lower performance, nevertheless it allows saving disk memory, as shown in the third column of Table \ref{tab:img_class_full}, thus it may be more appropriate for edge applications. On the other hand, processing color images with $n=3$ may bring higher accuracy even so it requires more parameters. Therefore, such a flexibility makes PHNNs adaptable to a large range of applications.
Likewise, PHResNets with $n=2$ gain considerable accuracy scores with respect to the real-valued corresponding models and, due to the larger number of parameters with respect to the PH model with $n=3$, sometimes outperform it too. Finally, the PHResNet with $n=4$ obtains the overall best accuracy in the largest experiment of this set. Indeed, considering a ResNet152 backbone on CIFAR100, our method exceeds the real-valued baseline by more than $4\%$. This is the empirical proof that, PHNNs well scale to large real-world problems by notably reducing the overall number of parameters. These results are summarized for ResNets and VGGs models on CIFAR10 in Fig.~\ref{fig:bubble}. The plot displays models accuracies against models parameters. The PH-based models, either ResNets or VGGs exceed their real and quaternion-valued baselines while consistently reduce the number of parameters. What is more, in Table \ref{tab:img_class_full}, we also report the memory required to store models checkpoints for inference. Our method crucially reduces the amount of disk memory demand with respect to the heavier real-valued model.

\begin{figure}
    \centering
    \includegraphics[width=\linewidth]{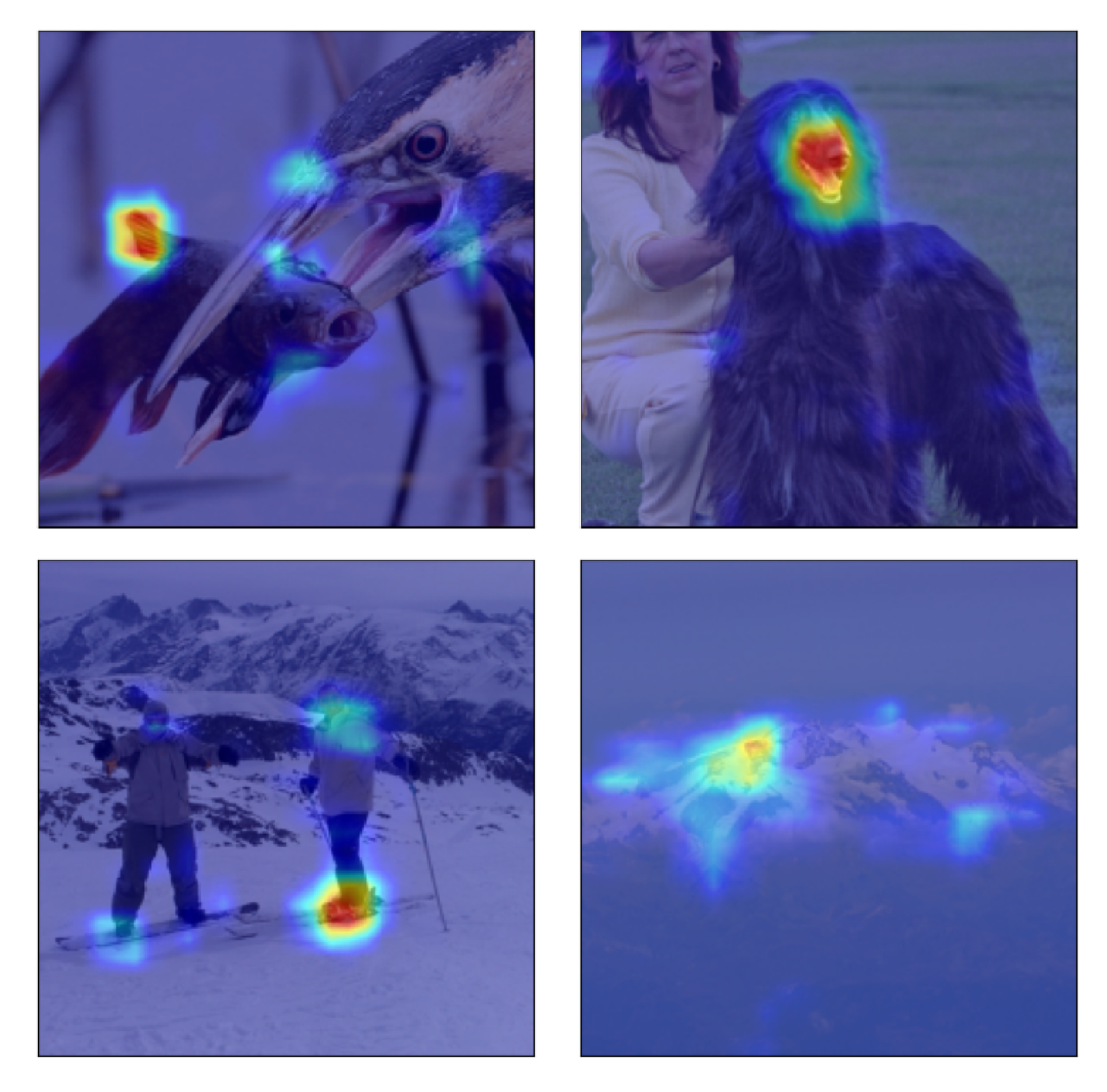}
    \caption{Grad-CAM visualization for the PHResNet50 $n=3$ on the ImageNet dataset.}
    \label{fig:gradcam}
\end{figure}

Further, we perform the image classifcation task on the ImageNet dataset. We compute the percentage of successes of ResNet-based models in each run for which we report the average accuracies in Table~\ref{tab:img_class_full}. As Fig.~\ref{fig:barplot} shows, the largest parcentage of successes is reached by the PHResNet with $n=3$ which has been demonstrated to be the most valuable choice for $n$ when dealing with RGB images. Therefore, we test the PHResNet with $n=3$ against the real-valued counterpart. 
Table \ref{tab:img_img} shows that the proposed method achieves comparable, and even slightly superior, performance than the real-valued baseline, while involving 66\% fewer parameters. Additionally, in Fig.\ref{fig:gradcam}, we provide Grad-CAM visualizations \cite{GradCAM2017ICCV} for a sample of predictions by our method in the ImageNet dataset to further prove the correct behavior of the PHResNet50 $n=3$ in this scenario. This proves the robustness of the proposed approach, which can be adopted and implemented in models at different scales.

\begin{table}[t]
\caption{ImageNet classification with real-valued baseline against our best model PH $n=3$. Our approach outperform the baseline while saving the $66\%$ of parameters.}
\label{tab:img_img}
\begin{center}
\begin{tabular}{llc}
\multicolumn{1}{c}{\bf Model} & \multicolumn{1}{c}{\bf Params} & \multicolumn{1}{c}{\bf ImageNet}\\
\hline \\
ResNet50         & 25.7M & 67.990 \\ 
PHResNet50 $n=3$ & 9.6M (-66\%) & \textbf{68.584} \\
\end{tabular}
\end{center}
\end{table}
%

\section{Experimental Evaluation on Sound Event Detection}
\label{sec:sed}

Sound event detection (SED) is the task of recognizing the sounds classes and at what temporal instances these sounds are active in an audio signal \cite{SED2021Mesaros}. We prove that the PHC layer is adaptable to $n$-dimensional input signals and, due to parameter reduction and hypercomplex algebra, is more performing in terms of efficiency and evaluation scores.

\subsection{Experimental Setup}

\begin{figure}[t]
    \centering
    \includegraphics[width=\linewidth]{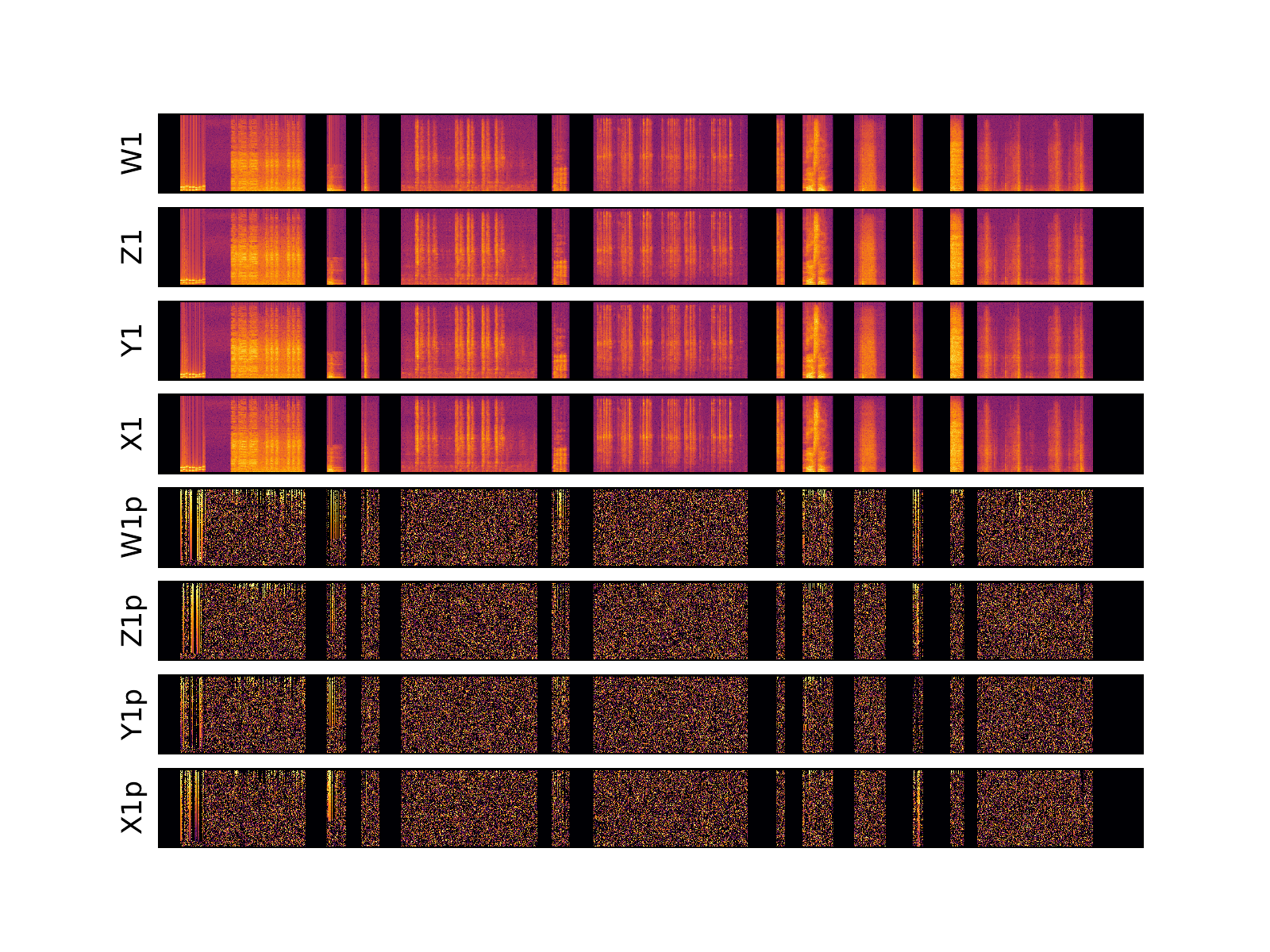}
    \caption{Sample spectrograms from L3DAS21 dataset recorded by one microphone with four capsules.The first four figures represent the magnitudes while the last four contain the corresponding phases information. The black sections represent silent instants.}
    \label{fig:l3das_dataset}
\end{figure}

For sound event detection models we consider the augmented version of the SELDnet \cite{Adavanne2019SoundEL, ComminielloICASSP2019a} which was proposed as baseline for of the L3DAS21 Challenge Task 2 \cite{guizzo2021l3das21} and we perform our experiments with the corresponding released dataset\footnote{L3DAS21 dataset and code are available at: \url{https://github.com/l3das/L3DAS21}.}. We consider as our baselines the SEDnet (without the localization part) and its quaternion counterpart. The L3DAS21 Task 2 dataset contains 15 hours of MSMP B-format Ambisonics audio recordings, divided in 900 1-minute-long data points sampled at a rate of $32$ kHz, where up to 3 acoustic events may overlap. The 14 sounds classes have been selected from the FSD50K dataset and are representative for an office sounds: \textit{computer keyboard, drawer open/close, cupboard open/close, finger snapping, keys jangling, knock, laughter, scissors, telephone, writing, chink and clink, printer, female speech, male speech}. In this dataset, the volume difference between the sounds is in the range $0$ and $20$ dB full scale (dBFS). Considering the array of two microphones $1, 2$, the channels order is [W1, Z1, Y1, X1, W2, Z2, Y2, X2], where W, X, Y, Z are the B-format ambisonics
channels if the phase (p) information is not considered. Whether we want to include also this information, the order will be [W1, Z1, Y1, X1, W1p, Z1p, Y1p, X1p, W2, Z2, Y2, X2, W2p, Z2p, Y2p, X2p] up to $16$ channels. In Fig.\ref{fig:l3das_dataset}, we show the $8$-channel input when considering one microphone and the phase information. Magnitudes and phases are normalized to be centered in $0$ with standard deviation $1$.

We perform experiments with multiple configurations of this dataset. We first test the recordings from one microphone considering the magnitudes only ($4$ channels input), then we test the networks with the signals recorded by two microphones and magnitudes only ($8$ channels input). The extracted features by the preprocessing are fed to the four-layer convolutional stack with $64,128,256,512$ filters, with batch normalization, ReLU activation, max pooling and dropout (probability $0.3$), with pooling sizes $(8,2),(8,2),(2,2),(1,1)$. The bidirectional GRU module has three layers, each with an hidden size of $256$. The tail is a four-layer fully connected classifier with $1024$ filters alternated by ReLUs and with a final dropout and a sigmoid activation function. The initial learning rate is set to $0.00001$. To be consistent with pre-existing literature metrics , we define True Positives as TP, False Positives as FP and False Negatives as FN. These are computed according to the detection metric \cite{guizzo2021l3das21}. Moreover, in order to compute the Error Rate (ER), we consider: $\text{S} = \min(\text{FN}, \text{FP})$, $\text{D} = \max(0, \text{FN}-\text{FP})$ and $\text{I} = \max(0, \text{FP}-\text{FN})$, as in \cite{Adavanne2019SoundEL, SED2021Mesaros}. Therefore, we consider:

\begin{equation*}
    \text{F\textsubscript{score}} = \frac{2 \text{TP}}{2\text{TP} + \text{FP} + \text{FN}},
\end{equation*}

\begin{equation*}
    \text{ER} = \frac{\text{S} + \text{D} + \text{I}}{\text{N}},
\end{equation*}

\noindent whereby $N$ is the total number of active sound event classes in the reference. The SED\textsubscript{score} is defined by:

\begin{equation*}
    \text{SED\textsubscript{score}} = \frac{\text{ER} + 1 - \text{F\textsubscript{score}}}{2}.
\end{equation*}

For ER and SED\textsubscript{score}, the lower scores, the better the performance, while for the F\textsubscript{score} higher values stand for better accuracy.

\subsection{Experimental Results}

\begin{table*}[t]
\caption{SEDnets results with one microphone ($4$ channels input). Scores are computed over three runs with different seeds and we report the mean. The proposed method wtih $n=2$ far exceeds the baselines in each metric considered.}
\label{tab:sed_4c}
\begin{center}
\begin{tabular}{llccccc}
\multicolumn{1}{c}{\bf Model} &\multicolumn{1}{c}{\bf Conv Params} &\multicolumn{1}{c}{\bf \text{F\textsubscript{score}} $\uparrow$} &\multicolumn{1}{c}{\bf ER $\downarrow$} &\multicolumn{1}{c}{\bf \text{SED\textsubscript{score}} $\downarrow$} &\multicolumn{1}{c}{\bf P $\uparrow$} &\multicolumn{1}{c}{\bf R $\uparrow$} \\
\hline \\

SEDnet         & 1.6M   & 0.637          & \underline{0.450}          & \underline{0.406}          & 0.756 & \underline{0.5505} \\
Quaternion SEDnet & 0.4M (-75\%) & 0.580          & 0.516          & 0.468          & 0.724          & 0.484 \\
PHSEDnet $n=2$ & 0.8M (-50\%) & \textbf{0.680} & \textbf{0.389} & \textbf{0.355} & \textbf{0.767} & \textbf{0.611} \\
PHSEDnet $n=4$ & 0.4M (-75\%) & \underline{0.638}          & 0.453          & 0.407          & \underline{0.765}    & 0.547\\
\end{tabular}
\end{center}
\end{table*}

\begin{table*}[t]
\caption{SEDnets results with two microphones ($8$ channels input). Scores are computed over three runs with different seeds and we report the mean. The PHSEDnet $n=2$ outperform the baselines. For training time (seconds/iteration) the mean and the standard deviation over one epoch is reported, for inference time we report the time required to perform an iteration on the validation set. PH-based models far exceed baselines both in training and inference time.}
\label{tab:sed_8c}
\begin{center}
\begin{tabular}{llccccccc}
\multicolumn{1}{c}{\bf Model} &\multicolumn{1}{c}{\bf Conv Params} &\multicolumn{1}{c}{\bf \text{F\textsubscript{score}} $\uparrow$} &\multicolumn{1}{c}{\bf ER $\downarrow$} &\multicolumn{1}{c}{\bf \text{SED\textsubscript{score}} $\downarrow$} &\multicolumn{1}{c}{\bf P $\uparrow$} &\multicolumn{1}{c}{\bf R $\uparrow$} &\multicolumn{1}{c}{\bf Time (T)} &\multicolumn{1}{c}{\bf Time (I)} \\
\hline \\

SEDnet         & 1.6M   & \underline{0.663} & \underline{0.428} & \underline{0.383} & \textbf{0.788} & \underline{0.572} & 1.242 $\pm$ 0.088 & 1.198 \\
Quaternion SEDnet & 0.4M (-75\%) & 0.559 & 0.556 & 0.499 & 0.754 & 0.444 & 1.308 $\pm$ 0.088 & 1.298 \\
PHSEDnet $n=2$ & 0.8M (-50\%) & \textbf{0.669} & \textbf{0.406} & \textbf{0.368} & \underline{0.767} & \textbf{0.594} & \textbf{1.091 $\pm$ 0.074} & \underline{1.085} \\
PHSEDnet $n=4$ & 0.4M (-75\%) & 0.638 & 0.433 & 0.397 & 0.729 & 0.567 & \textbf{1.091 $\pm$ 0.032} & \textbf{1.077} \\
PHSEDnet $n=8$ &  0.2M (-87\%) & 0.553 & 0.560 & 0.503 & 0.747 & 0.439 & \underline{1.142 $\pm$ 0.042} & 1.173 \\
\end{tabular}
\end{center}
\end{table*}

\begin{figure}[t]
    \centering
    \includegraphics[width=\linewidth]{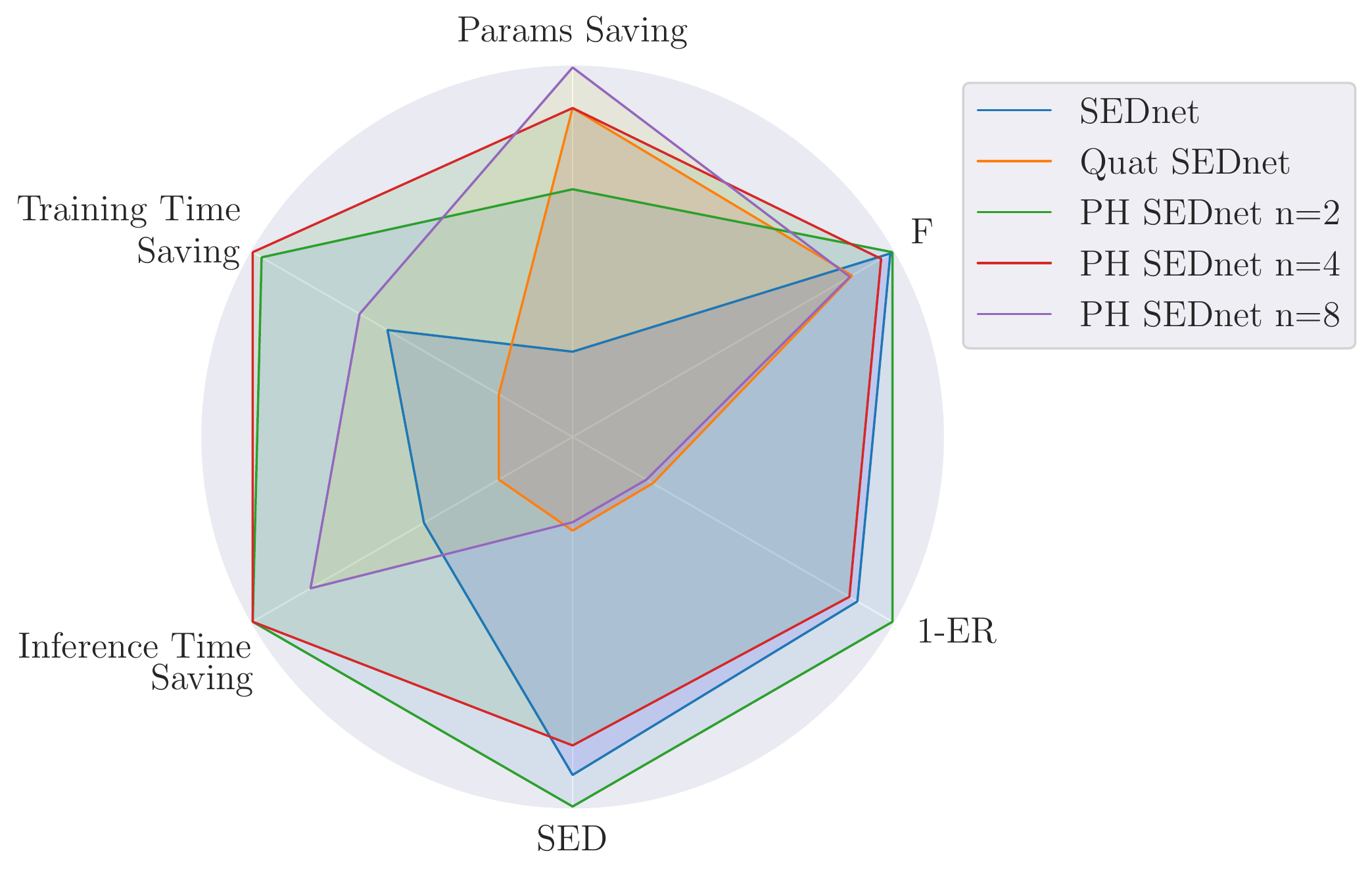}
    \caption{Radar plot for SEDnets results on L3DAS21 dataset with two microphones. The larger is the area, the better is the results. With the same computational time, PHC $n=2$ gains better scores with respect to PHC $n=4$ at a cost of more parameters. The real-valued SEDnet, although the discrete SED scores, has a high computational time demand as well the largest number of parameters.}
    \label{fig:radar}
\end{figure}

We investigate PHSEDnets in complex, quaternion and octonion domain with $n=2,4,8$ and train each network for $1000$ epochs with a batch size of $16$.
The proposed parameterized hypercomplex SEDnets distinctly outperform real and quaternion-valued baselines, as reported in Table \ref{tab:sed_4c} and Table \ref{tab:sed_8c}. Indeed, the PHSEDnet with $n=2$ gains the best results for each score and in both one and two microphone datasets, proving that the weights sharing due to the hypercomplex parameterization is able to capture more information regardless the lower number of parameters. It is interesting to note that the PHSEDnet $n=4$, which operates in the quaternion domain, achieves improved scores with respect to the Quaternion SEDnet that follows the rigid predefined algebra rules. Further, the malleability of PHC layers allows gaining comparable performance with respect to the quaternion baseline even so reducing convolutional parameters by $87\%$, just setting $n=8$. In Section \ref{ssubsec:sedresults}, we show additional experimental results of PH models able to save $94\%$ of convolutional parameters while operating in the sedonion domain by involving $n=16$.

Furthermore, PHSEDnets are more efficient in terms of time required for training and inference. Table \ref{tab:sed_8c} shows also that each tested version of the proposed method is faster regards as the real SEDnet and the quaternion one, both at training and at inference time. Time efficiency is crucial in audio applications where networks are usually trained for thousands of epochs and datasets are very large and require protracted computations.

Figure~\ref{fig:radar} summarises number of parameters, metrics scores and computational time in a radar plot from which it is clear that PHSEDnet $n=2$ gains the best scores and a large time saving at a cost of more parameters with respect to other versions but the real one. A good trade-off is brought by the PH model $n=4$ which further reduces the number of parameters at the cost of slightly worse SED$_\text{score}$ and ER. Moreover, the real-valued SEDnet is capable of obtaining fair scores while having the largest parameters amount and high computational time demanding.

\section{Ablation Studies}
\label{sec:abl}

\subsection{Less parameters do not lead to higher generalization}
\label{ssubsec:imageresults}


\begin{table}[]
\caption{Experiments on SVHN dataset with the smallest networks from each family, ResNet20 and VGG11, the latter with modified number of filters in order to be divided by each value of $n$ and FC layers in the closing classifier. We test also the PHNN with $n=1$ to replicate the real domain which outperform the real-valued ResNet20.}
\label{tab:img_class_app2}
\begin{center}
\begin{tabular}{llc}
\multicolumn{1}{c}{\bf Model} &\multicolumn{1}{c}{\bf Params} &\multicolumn{1}{c}{\bf SVHN} \\
\hline \\
ResNet20             & 0.27M         & 90.463  \\
Quaternion ResNet20  & 0.07M (-75\%) & 93.535  \\
PHResNet20 $n=1$   & 0.27M & \textbf{93.796}  \\
PHResNet20 $n=2$   & 0.14M (-50\%) & \underline{93.708}  \\
PHResNet20 $n=4$   & 0.07M (-75\%)  & 93.669  \\
\hline
VGG11            & 13.8M        & 93.488  \\
Quaternion VGG11 & 3.9M (-71\%) & 92.888  \\
PHVGG11 $n=2$  & 7.2M (-48\%) & \textbf{93.958}  \\
PHVGG11 $n=3$  & 5.0M (-64\%) & 93.804  \\
PHVGG11 $n=4$  & 3.9M (-71\%) & \underline{93.919}  \\
\end{tabular}
\end{center}
\end{table}

\begin{table}[]
\caption{The first lines report VGG16 results with real-valued classifier for quaternion and PHNNs. Extension of Table \ref{tab:img_class}. Additional experiments with ResNet56 and ResNet110, the latter with modified number of filters in order to be divided by each value of $n$. Accuracy score is the mean over three runs with different seeds.}
\label{tab:img_class_app3}
\begin{center}
\begin{tabular}{llcc}
\multicolumn{1}{c}{\bf Model} &\multicolumn{1}{c}{\bf Params} &\multicolumn{1}{c}{\bf SVHN} &\multicolumn{1}{c}{\bf CIFAR10} \\
\hline \\
Quaternion VGG16 & 4.2M (-72\%) & 94.086 & 84.126 \\
PHVGG16 $n=2$  & 7.9M (-62\%) & \textbf{94.885} & \textbf{86.147} \\
PHVGG16 $n=4$  & 4.2M (-72\%) & \underline{94.562} & \underline{85.710} \\ \hline
ResNet56             & 0.9M         & \underline{94.116} & \textbf{83.700}  \\
Quaternion ResNet56  & 0.2M (-75\%) & 93.664 & 81.687  \\
PHResNet56 $n=2$   & 0.4M (-50\%) & 93.722 & \underline{83.413}  \\
PHResNet56 $n=4$   & 0.2 (-75\%)  & \textbf{94.122} & 82.720  \\
\hline
ResNet110            & 16.7M        & 93.461 & 84.810  \\
Quaternion ResNet110 & 4.2M (-75\%) & 92.788 & 83.920  \\
PHResNet110 $n=2$  & 8.4M (-50\%) & 93.746 & 83.220  \\
PHResNet110 $n=3$  & 5.6M (-66\%) & \underline{94.712} & \underline{85.200}  \\
PHResNet110 $n=4$  & 4.2M (-75\%) & \textbf{94.885} & \textbf{85.280}  \\
\end{tabular}
\end{center}
\end{table}

In the following, we demonstrate that higher accuracies achieved by our method are not caused by the parameter reduction which may lead to more generalization. To this end, we perform multiple experiments. First, we test lighter ResNets that were originally built for the CIFAR10 dataset \cite{Resnet2016}: ResNet20, ResNet56 and ResNet110. Second, we consider also the smallest VGG network, that is the VGG11 which has $14$M parameters. Finally, we perform experiments on SVHN, CIFAR10 and CIFAR100 with the larger ResNet18, ResNet50 and ResNet152 reducing the number of filters by $75\%$ so to have the same number of parameters of quaternion and PHNN with $n=4$ counterparts.

Table \ref{tab:img_class_app2} reports experiments with ResNet20 where we test also $n=1$ to replicate the real-valued model, outperforming it. Experiments with VGG11 with modified number of filters in order to be divided by each value of $n$ is also reported in the same table. Finally, in Table \ref{tab:img_class_app3} we report experiments on SVHN and CIFAR10 with ResNet56 and ResNet110, the latter with modified number of filters.
PH models gain good performance in each test we conduct while reducing the amount of free parameters. Indeed, the PHResNet20s gain almost $94\%$ of accuracy on the SVHN dataset involving just $70$k parameters.

\begin{table}[t]
\caption{Real-Valued ResNets with convolutional filters reduced by $75\%$, denoted by (s). Full models exceeds reduced versions in each of the experiment, proving that a smaller number of parameters do not lead to higher generalization capabilities.}
\label{tab:red_real}
\begin{center}
\begin{tabular}{llccc}
\multicolumn{1}{c}{\bf Model} &\multicolumn{1}{c}{\bf Params} &\multicolumn{1}{c}{\bf SVHN} &\multicolumn{1}{c}{\bf CIFAR10} &\multicolumn{1}{c}{\bf CIFAR100} \\
\hline \\
ResNet18            & 10.1M        & \textbf{93.992} & \textbf{89.543} & \textbf{62.634}  \\
ResNet18 (s)  & 2.7M (-75\%) & 93.842 & 88.310 & 59.590  \\
\hline
ResNet50            & 22.5M        & \textbf{94.546} & \textbf{89.630} & \textbf{65.514}  \\
ResNet50 (s)  & 5.7M (-75\%) & 93.915 & 89.370 & 62.450  \\
\hline
ResNet152           & 52.6M        & \textbf{94.625} & \textbf{89.580} & \textbf{62.053}  \\
ResNet152 (s) & 13.2M (-75\%)& 94.400 & 89.001 & 60.850  \\
\end{tabular}
\end{center}
\end{table}

Finally, in order to further remove the hypothesis that smaller number of neural parameters leads to higher generalization capabilities, we perform experiments with real-valued baselines with a number of parameters reduced by $75\%$. Table \ref{tab:red_real} shows that reducing the number of filters downgrades the performance and thus it is not sufficient to improve the generalization capabilities of a model. We do not include standard deviations for values in the ablation studies as the values are similar to the previous examples so we aim at favoring paper readability.

\subsection{Push the hyperparameter $n$ up to $16$}
\label{ssubsec:sedresults}

In the following, we perform additional experiments for the sound event detection task. We conduct a test considering two microphones and the phase information, so to have an input with $16$ channels. For this purposes, we consider as baseline the quaternion model and PHNNs with $n=4,8,16$ so to test higher order domains. Quaternion and PHSEDnet with $n=4$ manage the $16$ channels by grouping them in four components, thus assembling them in $4$ channels: one channel containing the magnitudes of the first microphone, one channel the phases of the same microphone, and so on. Therefore, the details coming from the magnitudes, which are the most important for sound event detection, are grouped together without properly exploiting this information. On the contrary, employing PHC layers allows the model to process information without roughly grouping channels while instead leveraging every information by easily setting $n$ equal to the number of channels, that is in this case $16$. From Table~\ref{tab:sed_app}, it is clear that employing a $4$-channel model such as Quaternion or PHC with $n=4$ does not lead to higher performance, despite the higher number of parameters. Indeed, the best scores are obtained with PHC models involving $n=8$ and $n=16$ that are able to grasp information from each channel.

\begin{table*}[t]
\caption{SED results with two microphone: magnitudes and phases (16 channels input). We test higher order hypercomplex domains up to sedonions by setting $n=16$. Although the incredible reduction of the number of parameters with respect to the real-valued baseline in Table \ref{tab:sed_8c}, the PHNN with $n=16$ still has comparable performance with other models. Furthermore, the PHSEDnet with $n=8$ outperform also the quaternion baseline which has more degrees of freedom.}
\label{tab:sed_app}
\begin{center}
\begin{tabular}{lcccccc}
\multicolumn{1}{c}{\bf Model} &\multicolumn{1}{c}{\bf Conv Params} &\multicolumn{1}{c}{\bf \text{F\textsubscript{score}} $\uparrow$} &\multicolumn{1}{c}{\bf ER $\downarrow$} &\multicolumn{1}{c}{\bf \text{SED\textsubscript{score}} $\downarrow$} &\multicolumn{1}{c}{\bf P $\uparrow$} &\multicolumn{1}{c}{\bf R $\uparrow$} \\

\hline \\

Quaternion SEDnet & 0.4M (-75\%) & 0.580          & 0.480          & 0.450          & 0.655          & 0.520 \\
PHSEDnet $n=4$ & 0.4M (-75\%) & 0.585 & \underline{0.470} & \underline{0.443} & 0.653          & \underline{0.530} \\
PHSEDnet $n=8$ & 0.2M (-87\%) & \textbf{0.607}          & \textbf{0.466}          & \textbf{0.430}          & \underline{0.702}          & \textbf{0.534} \\
PHSEDnet $n=16$ & 0.1M (-94\%) & \underline{0.588}          & 0.509          & 0.461          & \textbf{0.734}          & 0.491 \\

\end{tabular}
\end{center}
\end{table*}

\section{Conclusion}
\label{sec:conc}
In this paper, we introduce a parameterized hypercomplex convolutional (PHC) layer which grasps the convolution rule directly from data and can operate in any domain from $1$D to $n$D, regardless the algebra regulations are preset. The proposed approach reduces the convolution parameters to $1/n$ with respect to real-valued counterparts and allows capturing internal latent relations thanks to parameter sharing among input dimensions. Employing this method, jointly with the one in \cite{Zhang2021PHM}, we devise the family of parameterized hypercomplex neural networks (PHNNs), a set of lightweight and efficient neural models exploiting hypercomplex algebra properties for increased performance and high flexibility.
We show our method is flexible to operate in different fields of application by performing experiments with images and audio signals. We also prove the malleability and the robustness of our approach to learn convolution rules in any domain by setting different values for the hyperparameter $n$ from $2$ to $16$.

\subsection*{CO2 Emission Related to Experiments}

Experiments were conducted using a private infrastructure, which has a carbon efficiency of 0.445 kgCO$_2$eq/kWh. A cumulative of 2000 hours of computation was performed on hardware of type Tesla V100-SXM2-32GB (TDP of 300W). Total emissions are estimated to be 267 kgCO$_2$eq of which 0 percents were directly offset. Estimations were conducted using the \href{https://mlco2.github.io/impact#compute}{MachineLearning Impact calculator} presented in \cite{lacoste2019quantifying}.

More in detail, considering an experiment for the sound event detection (SED) task, according to Table \ref{tab:sed_8c}, the real-valued baseline requires approximately 20 hours for training and validation, with a corresponding carbon emissions of $2.71$ kgCO$_2$eq. Conversely, the proposed PH model takes approximately $17$ hours with a reduction of carbon emissions of $16\%$, being $2.28$ kgCO$_2$eq.

In conclusion, we believe that the improved efficiency of our method with respect to standard models may be a little step towards reducing carbon emissions.

\bibliographystyle{IEEEtran}
\bibliography{PHCbib}


\end{document}